\documentclass{article} 
\usepackage{amssymb}
\usepackage{amsmath}
\usepackage{amsthm}
\usepackage[a4paper]{geometry}
\usepackage[round]{natbib}
 \usepackage[colorlinks=true, linkcolor=blue]{hyperref}

\usepackage{amssymb}
\usepackage{amsmath,amscd}
\usepackage{amsthm}
\usepackage{cancel}     \usepackage{centernot}  \usepackage{graphicx}   \usepackage{soul}       \usepackage{color}      \usepackage{enumitem}   \usepackage{commath}    \usepackage{algorithm}  \usepackage{subcaption} \usepackage{booktabs}   \usepackage{multirow} 

\newtheorem{thm}{Theorem}[section]
\newtheorem{lemma}[thm]{Lemma}
\newtheorem{cor}[thm]{Corollary}

\newtheorem*{thm*}{Theorem}
\newtheorem*{lemma*}{Lemma}
\newtheorem*{cor*}{Corollary}
\newtheorem*{prop*}{Proposition}
\newtheorem*{conjecture*}{Conjecture}

\theoremstyle{definition}

\newtheorem*{defn*}{Definition}
\theoremstyle{definition}

\theoremstyle{definition}

\theoremstyle{remark}
\newtheorem*{ex*}{Example}
\theoremstyle{definition}

\theoremstyle{definition}
\newtheorem*{assm*}{Assumption}
\theoremstyle{remark}
\newtheorem{remark}{Remark}[section]
\theoremstyle{remark}
\newtheorem*{remark*}{Remark}

\DeclareFontFamily{U}{mathx}{\hyphenchar\font45}
\DeclareFontShape{U}{mathx}{m}{n}{
      <5> <6> <7> <8> <9> <10> gen * mathx
      <10.95> mathx10 <12> <14.4> <17.28> <20.74> <24.88> mathx12
      }{}
\DeclareSymbolFont{mathx}{U}{mathx}{m}{n}
\DeclareMathSymbol{\intop}  {1}{mathx}{"B3}

\DeclareFontFamily{U}{mathx}{\hyphenchar\font45}
\DeclareFontShape{U}{mathx}{m}{n}{
      <5> <6> <7> <8> <9> <10>
      <10.95> <12> <14.4> <17.28> <20.74> <24.88>
      mathx10
      }{}
\DeclareSymbolFont{mathx}{U}{mathx}{m}{n}
\DeclareFontSubstitution{U}{mathx}{m}{n}
\DeclareMathAccent{\widecheck}{0}{mathx}{"71}
\DeclareMathAccent{\wideparen}{0}{mathx}{"75}

\newcommand\indep{\independent}
\newcommand\independent{\protect\mathpalette{\protect\independenT}{\perp}}
\def\independenT#1#2{\mathrel{\rlap{$#1#2$}\mkern4mu{#1#2}}}

\newcommand{\wh}{\widehat}

\let\temp\phi
\let\phi\varphi
\let\varphi\temp

\newcommand{\pr}{\mathbb{P}}
\newcommand{\R}{\mathbb{R}}

\newcommand{\E}{\mathbb{E}}

            \newcommand{\given}{\,|\,}

\renewcommand{\norm}[1]{\Vert#1\Vert}

\newcommand{\les}{\lesssim}

\DeclareMathOperator{\Eig}{Eig}
\DeclareMathOperator{\col}{col}
\DeclareMathOperator{\diagonal}{diagonal}

\DeclareMathOperator{\cov}{cov}

\DeclareMathOperator{\rank}{rank}

\DeclareMathOperator{\diag}{diag}

\newcommand{\gr}{\mathsf{G}}
\newcommand{\ver}{\mathsf{V}}
\newcommand{\edg}{\mathsf{E}}

\newcommand{\FORALL}{\text{ for all }}

\DeclareMathOperator{\poly}{poly}
\newcommand{\dimz}{d}

\title{Beyond identifiability: Learning causal representations\\with few environments and finite samples}
\author{Inbeom Lee \and Tongtong Jin \and Bryon Aragam}

\begin{document}
\maketitle

\begin{abstract}
We provide explicit, finite-sample guarantees for learning causal representations from data with a sublinear number of environments. Causal representation learning seeks to provide a rigourous foundation for the general representation learning problem by bridging causal models with latent factor models in order to learn interpretable representations with causal semantics.
Despite a blossoming theory of identifiability in causal representation learning, estimation and finite-sample bounds are less well understood. 
We show that causal representations can be learned with only a logarithmic number of unknown, multi-node interventions, and that the intervention targets need not be carefully designed in advance.
Through a careful perturbation analysis, we provide a new analysis of this problem that guarantees consistent recovery of (a) the latent causal graph, (b) the mixing matrix and representations, and (c) \emph{unknown} intervention targets. 
\end{abstract}

\section{Introduction}
The impressive performance of generative models is due in large part to their ability to learn internal ``representations'' from unstructured datasets. Unfortunately, owing to the complexity of modern generative models, probing these representations is difficult if not impossible. Causal representation learning \citep[CRL,][]{scholkopf2021towards} is an emerging discipline that probes the statistical foundations of representation learning from a causal perspective: Models should learn abstract representations that come paired with causal semantics and interpretability ``baked in''. 
As a result, models generalize better out-of-distribution \citep{fan2024environment,peters2016causal}, learn abstract concepts \citep{taeb2022provable,rajendran2024causal}, and can be probed \citep{de2025interpretable,jin2025discovering}.
By exploiting causal models and assumptions, it is now known that identifiability of the latent representations is possible across a variety of settings and applications. For a statistical perspective on this problem, see \citet{moran2026towards}.

Despite this rich and developing theory of identifiability in CRL, even basic statistical properties such as consistency and finite-sample estimation are poorly understood. Surprisingly, this remains the case even after restricting to simple families such as linear models. On the other hand, this is unsurprising given the well-known difficulties with estimation in latent factor models \citep{drton2009likelihood,bai2012statistical,auddy2025large} and causal models \citep{robins2003uniform,uhler2013geometry}. Indeed, either of these problems on its own presents substantial challenges, so combining these problems only makes things that much more difficult. Moreover, it is not clear to what extent various assumptions are necessary, or if stronger assumptions are needed to translate existing identifiability results into estimators with appropriate guarantees.

In this paper, we contribute to closing this gap and provide finite-sample, nonasymptotic guarantees for CRL in high-dimensional, linear factor models. 
Achieving finite-sample rates in this setting involves several novel techniques in order to bridge the dual complications that arise from combining causality and factor models. 
Unlike traditional factor models that either assume uncorrelated factors or that impose sparsity conditions on the factor structure, and unlike traditional causal models that often assume faithfulness,
our results require neither.
This takes us a step beyond traditional factor models by integrating techniques from causality and model selection, which fundamentally alters the statistical setting that we are in.
Although we focus on the linear setting, our results represent a necessary and important first step towards the general case of principled representation learning in nonlinear generative models.

As in classical models, without additional assumptions neither the latent factors (a.k.a. representations) nor their causal relationships are identifiable. A key driver of recent developments in CRL is the observation that data arising from multiple environments can help break the identifiability barrier. 
A common type of environment arises from causal interventions in the latent space:
In vision applications, these may be concept edits or style manipulation arising from data augmentation. In biological applications, these may be CRISPR knockouts or drug administration. More generally, any proxy intervention where the observables are indirectly perturbed.
For our results we will assume we have data arising from multiple environments, where each environment arises from latent, unknown interventions on these factors.

This gives rise to a natural tradeoff: Additional environments give stronger identifiability, but this comes at the cost of more expensive data collection. The question that arises then is: \emph{What is the fewest number of environments $K$ necessary?} 
With interventional environments, it is known that identifying causal models is possible with $K=o(\dimz)$ environments \citep{eberhardt2005number}, where $\dimz$ is the number of causal variables (in our setting, this will be the number of latent factors, also known as the latent dimension). 
The natural question that emerges is whether or not this bound is achievable for learning causal representations, and is closely related to important problems in out-of-distribution generalization \citep{peters2016causal,rosenfeld2020risks,chen2022iterative,fan2024environment,gu2025causality}. When restricted to \emph{single-target} interventions, a known necessary condition \citep{squires2023linear} is that $K=\Omega(\dimz)$. 
Thus, improving existing results to require only \emph{sublinear} environments necessitates consideration of \emph{multi-target} interventions,
where it is known that identification is possible with $K=O(\log\dimz)$ interventions \citep{eberhardt2005number} when the variables are \emph{observed} and the targets in each environment are \emph{known}. 
This leaves open the problem of identifying \emph{latent} causal models (as in CRL) with \emph{unknown} intervention targets all with only a logarithmic number of environments. 
While there have been recent advances on identifiability using multi-target interventions, translating these results into practical estimators with finite-sample guarantees has remained elusive.
One of the main contributions of this paper is to propose a new estimator that requires only logarithmic environments in this more challenging setting with \emph{unknown}, \emph{latent} targets and to characterize its finite-sample behaviour. 

We adopt the usual set-up of linear CRL: There is a latent (unobserved) causal model over the latent causal representations $Z$, and we observe $X=BZ$, where $X\in\R^p$ is potentially much higher dimensional than $Z\in\R^d$, i.e. $p\gg d$. The latent causal model over $Z$ is given by an unknown linear structural equation model (SEM) over $Z$. Taken together, the joint model between $X$ and $Z$ can be written as: 
\begin{align}
\label{eq:model}
\begin{aligned}
    X &= BZ, \quad B\in\R^{p\times d},
\\
    Z &= A^{T}Z + \nu,
    \quad A\in\R^{d\times d},
    \quad \nu_i \indep \nu_j.
\end{aligned}
\end{align}
The matrix $A$ defines a linear structural equation model over $Z$ with a corresponding directed acyclic graph (DAG) $\gr$---also known as the latent causal graph---that will be our primary interest, along with the causal representations $Z$. We do not assume $\gr$ is known, and we seek to learn it from data. We also estimate the mixing matrix $B$ that decodes how representations are transformed into the observations $X$. For this reason the map $Z\mapsto BZ$ is sometimes called a ``decoder''.

\paragraph{Notation}
For an integer $K\ge 1$, we use the set notation $[K]:=\{1,\dots,K\}$. We denote the cardinality of a set $S$ by $|S|$. We write $\col(W)$ for the column space of a matrix $W$, $\rank(W)$ for its rank, and $P_{W}$ for the orthogonal projector onto $\col(W)$. The dimension of a linear space $\mathcal{U}$ is denoted by $\dim(\mathcal{U})$. The $i$-th row, $j$-th column, and $(i,j)$-th element of a matrix $W$ are denoted by $W_{[i,\cdot]}$, $W_{[\cdot, j]}$, and $W_{[i,j]}$, respectively.  $\norm{W}_2$ and $\norm{W}_F$ denote the spectral and Frobenius norms, respectively, and $W^\dagger$ denotes the Moore-Penrose pseudoinverse. We denote the minimum non-zero and maximum singular values as  $\sigma_{\min}(W), \sigma_{\max}(W)$, respectively, the minimum non-zero and maximum eigenvalue as $\lambda^{+}_{\min}(W),~ \lambda_{\max}(W)$, respectively, and the $j$-th largest eigenvalue as $\lambda_j(W)$. For any two sequences $a_n$ and $b_n$, we write $a_n \lesssim b_n$ if there exists some fixed positive constant $C$ such that $a_n \leq Cb_n$. We also use the following notation to refer to the maximum and minimum: $a \vee b = \max(a,b)$, $a \wedge b = \min(a,b)$.

\subsection{Related work}\label{sec.related.work}
Our results relate most closely to parallel lines of work on causal representation learning and latent factor models. Below we discuss our contributions in the context of this related work.

\subsubsection{Causal representation learning}

Causal representation learning \citep{scholkopf2021towards} has its origins in the structural equation modeling and factor analysis literature \citep{bollen1989structural}; the specific linear causal model considered here was first studied in \citet{silva2006learning}. 
There is also a close relationship to the ICA literature; see \citet{hyvarinen2016unsupervised,khemakhem2020variational} for important contributions on the role multi-environment data can play in identifying latent representations.
We refer interested readers to these papers for additional historical references on the broader identifiability problem. In recent years, the identifiability theory of both linear and nonlinear CRL models has blossomed. Although we do not intend a detailed review, we discuss some of the most closely related works below and defer to \citet{moran2026towards} for a more in-depth discussion.

For causal identifiability, CRL often exploits the additional information afforded by multiple environments such as latent interventions that are widely available in applications. With interventions, several key issues arise: 
\begin{enumerate}
    \item Are the interventions latent or observed?
    \item Are the intervention targets known or unknown?
    \item How many interventions are required? 
\end{enumerate}
Most identifiability results assume latent, single-node interventions on every node \citep[starting with][]{ahuja2022interventional,squires2023linear,buchholz2023learning,von2023nonparametric,jin2023learning}. More closely related to our setting, several identifiability results use multi-target interventions \citep{lippe2022intervention,lippe2023biscuit,bing2024identifying,talon2024towards,ahuja2024multi,li2024disentangled,ng2025causal}. An advantage of multi-node interventions is that one only needs (in principle) a \emph{sublinear} number of environments. A key contribution of our work is to obtain precise, finite-sample error bounds in exactly this setting with sublinear environments. Moreover, we accomplish this without assuming the interventions targets are known, as commonly assumed \citep{lippe2022intervention,talon2024towards,li2024causal}. 

By comparison, the extension of these identifiability results to provably consistent estimators has been slower, arguably due to the well-known complications in learning both causal models and latent factor models. To the best of our knowledge, the only other work that provides error bounds is \citet{acarturk2024sample}, which establishes an $O(n^{-1/4})$ rate in the same noiseless, linear mixing setting of $X=BZ$ with unknown single-node interventions. As a result, this result necessarily requires $\Omega(\dimz)$ environments due to known lower bounds \citep{squires2023linear}. 
These results also rely on the rank-one fingerprint left by single-node interventions, a strategy which breaks down with multiple targets since source nodes can become mixed with non-source nodes and the resulting rank structure becomes more intricate.
Other related results include \citet{buchholz2025robustness}, which considers approximate identifiability in nonlinear ICA, and \citet{fokkema2025sample}, which simplifies the analysis by assuming that the decoder is known \emph{a priori}. By contrast, accounting for the ill-conditioning that arises from an unknown decoder that must be learned from data is a major hurdle in our analysis.

\subsubsection{Factor models}
Classical works have shown that the factor loading matrix $B$ (also called the mixing matrix or decoder) is not identifiable without extra conditions on the structure of $B$ or $\Sigma_Z$ \citep{anderson1956statistical,lawley1962factor}. Many assumptions have been proposed, each having different tradeoffs between assumptions on $B$ and $\Sigma_Z$, with some notable ones being (i) restricting the latent variables to be uncorrelated, (ii) requiring $B$ to have a $\dimz \times \dimz$ lower triangular matrix as a submatrix, and (iii) imposing a low-rank condition and fixing specific entries to be zero in $B$ \citep{joreskog1979advances,peeters2012rotational}. Our causal setting precludes type (i) assumptions.

One of the most common assumptions is the pure child assumption \citep{donoho2003does,silva2006learning,arora2013practical,bing2020optimal}, which is a strengthening of (ii). Several recent works invoke this pure child assumption \citep{bing2020adaptive,bunea2020model,bing2022inference,kim2023structure}, 
including several CRL identification results \citep{moran2022identifiable,huang2022latent,chen2024learning} and works involving a relaxation of pure children to a so-called \emph{subset condition} \citep{kivva2021learning,zheng2022identifiability,jiang2023learning}.

As our method is applicable to high-dimensional data, it naturally has ties to the \emph{high dimensional} factor analysis literature \citep{fan2008high,fan2011high,bai2012statistical,fan2013large,bing2022inference}, which encompasses works that take advantage of the latent factor structure in high-dimensional tasks such as covariance estimation \citep{fan2013large}, clustering \citep{bunea2020model}, and topic modeling \citep{bing2020adaptive}. 
Other related problems include ICA and low-rank matrix factorization, both of which avoid the causal framing as discussed previously. See \citet{montanari2024fundamental,auddy2025large,zhou2025deflated}.

The approach to identifying $B$ and the latent causal structure through multiple environments provides an alternative that relaxes the constraints on the latent structure. While the aforementioned statistical works provided sharp characterizations for finite sample recovery, they are less concerned with the implications the structural constraints had besides allowing identifiability. On the other hand, while CRL is interested in investigating the latent structure itself, less emphasis has been placed on finite sample results. Our work finds itself at the intersection of these two fields, and we combine the best of both worlds by allowing the latent structure to be unconstrained while also providing a likelihood-free estimator with finite-sample guarantees.

\section{Model}

In this section, we first introduce our model which includes the data generating mechanism and the intervention mechanism before presenting an identifiability result in Theorem \ref{thm.2.0}.

\subsection{Model}\label{sec.model}

Recall the basic model \eqref{eq:model}. The latent SEM parameters $A=(a_{uv})$ define a latent causal graph $\gr=\gr(A)=(\ver,\edg)$ in the obvious way: $\ver=Z$ and $(u,v)\in\edg$ if and only if $a_{uv}\ne0$. We do not impose specific distributional assumptions on the noise $\nu$; it is allowed to be non-Gaussian, Gaussian, or mixed. The mixing matrix $B$ is assumed to have full column rank, but otherwise no sparsity assumptions are imposed.
Our primary interest is to recover $\gr$ and the representations $Z$ from observations on $X$. 
Since $B$ is full rank, $Z$ can be easily recovered from $B$. Thus we will emphasize estimation of $B$ and $\gr$ in the sequel.

Without additional assumptions, neither $\gr$ nor $Z$ are identifiable. To rescue identifiability we assume we are given additional observations of $X$ arising from different environments in the form of latent interventions on the latent causal model. We intervene on both the latent factors and the noise, indexed by $k$ and $\ell$, respectively. We consider the general setting of multiple unknown intervention targets, i.e. we allow any number of nodes to be intervened on, and we denote the set of nodes that are intervened on in the $k$-th environment as $I(k)\subset[\dimz]$. Formally, we denote $X^{(k),\ell}$, $Z^{(k),\ell}$, $ \nu^{(k),\ell}$ to be the counterparts of the aforementioned $X,Z,\nu$ 
where nodes $I(k)$ are subject to intervention (for $k=0,1,\ldots,K$, with $k=0$ denoting the observational environment)
and the scale of the noise is shifted between $\ell=1$ and $\ell=2$.
It follows that our model can be expressed as: 
\begin{align}
    Z^{(k),\ell} ~&=~ [A^{(k)}]^{T}Z^{(k),\ell} + \nu^{(k),\ell}, 
    \quad \nu^{(k),\ell}_i\indep \nu^{(k),\ell}_j, \label{eq.linSEM} \\
    X^{(k),\ell} ~&=~
    BZ^{(k),\ell} 
    \label{eq.linmix}
\end{align}
where $A^{(k)} \in \R^{\dimz \times \dimz}$ represents the latent DAG in the $k$-th environment after intervention. 
We intervene on the nodes $I(k)$ by zero-ing out the corresponding columns in $A^{(k)}$ and diagonal entries in $\Sigma_\nu^{(k),\ell}:=\cov(\nu^{(k),\ell})$.
Without loss of generality we assume that all the latent factors $Z$ are centered, and thus that $X$ is centered. Aside from subgaussianity of $X$ in the finite-sample results, we will not impose any further distributional assumptions on $\nu$, $Z$. In particular, although linear, we do not assume Gaussianity in the noise, the latents, or the observations (see Remark~\ref{rem:errordist}).

We assume a standard diversity and coverage condition on the intervention design that stems from well-known necessary conditions~\citep{eberhardt2005number,hyttinen2013experiment}:

\begin{enumerate}[label=(A\arabic*)]
    \item\label{assum.PSSS.0}
(a) $K\les\log d$, (b) There is an observational environment $k=0$, and
    (c) For every pair $j_1 \neq j_2$, there exists some $k_1, k_2\in \{1,...,K\}$ such that $Z_{j_1}$ is intervened on in environment $k_1$ while $Z_{j_2}$ isn't and $Z_{j_2}$ is intervened on in environment $k_2$ while $Z_{j_1}$ isn't. 
\end{enumerate}
\noindent
The condition that $K\les\log d$ is not necessary; in our technical results the dependence on $K$ is made explicit in the rates. We emphasize the dependence between $K$ and $\dimz$ here to emphasize that we are most interested in sublinear environments, which is the minimum the number of environments possible. 

Condition \ref{assum.PSSS.0} is the conventional \emph{strongly separating systems} assumption \citep{hyttinen2013experiment} combined with an observational environment. 
Condition \ref{assum.PSSS.0} is practically relevant as it is applicable to combinatorial/multiplex CRISPR data, where multiple genes are simultaneously perturbed in each environment \citep{shen2017combinatorial, horlbeck2018mapping}. 

We also impose a technical condition on the noise interventions: \begin{enumerate}[label=(A\arabic*),start=2]
    \item\label{assum.dnvr.0}
    The two noise covariance matrices, $\Sigma_{\nu}^{(0),1}, \Sigma_{\nu}^{(0),2}$ where $\Sigma_{\nu}^{(0),1} = \diag(\sigma_{1,1}, ~...~, \sigma_{\dimz,1})$, $\Sigma_{\nu}^{(0),2} = \diag(\sigma_{1,2}, ~...~, \sigma_{\dimz,2})$ satisfy 
    \begin{align*}
        \frac{\sigma_{j,1}}{\sigma_{j,2}} ~\neq~ \frac{\sigma_{j',1}}{\sigma_{j',2}}~\FORALL j \neq j' 
    \end{align*}
\end{enumerate}

\noindent
Similar conditions on the noise have been used previously for various identifiability results.
For example, \citet{rothenhausler2015backshift} applied a similar noise variance ratio condition to flow cytometry data (on biomarker levels in single cells) in multiple environments corresponding to shift interventions. 
\citet{kivva2022identifiability} apply this condition to identify generative models comprised of ReLU networks.

\subsection{Identifiability}

We now present an identifiability result that ensures the estimation problem is well-defined and that provides the foundation for our finite-sample results:
\begin{thm} \label{thm.2.0}
    $\mathrm{(Identifiability ~ with ~ unknown~intervention~environments)}$ Under the model (\ref{eq.linSEM}-\ref{eq.linmix}) and \ref{assum.PSSS.0}-\ref{assum.dnvr.0}, the following can be identified from $K$ unknown, multi-node intervention environments:
\begin{enumerate}[label=(\alph*)]
        \item the latent causal graph $\gr(A)$ (up to label permutation),
        \item the causal representations $Z$ (up to scale and label permutation),
        \item the decoder $B$ (up to scale and label permutation),
        \item the intervention targets $I(k)$ for each $k$ (up to label permutation).
    \end{enumerate}
\end{thm}
\noindent
We also include the following corollary to highlight that sublinear environments is indeed sufficient for identifiability:
\begin{cor} \label{cor.log.0}
    $K\asymp\log d$ environments suffice to identify the latent causal graph, causal representations, decoder, and intervention targets.
\end{cor}
\noindent
This is information-theoretically optimal in the sense that, even if all the latent factors were fully observed and all the intervention targets were known, then $\Omega(\log d)$ interventions are known to be necessary \citep{eberhardt2005number}.
 Moreover, since the ambient dimension $p$ is typically much larger than the embedding dimension $d$, we also have
\begin{align*}
    K \ll d \ll p.
\end{align*}
Although not always required in our technical results, this reflects the general setting that we are most interested in.

\section{Estimation}\label{sec.stat}

Our main goal is to move beyond identifiability results such as Theorem~\ref{thm.2.0} towards a precise characterization of the typical finite-sample estimation behaviour in this model. It is worth pausing to consider the available strategies.

For example, if we were impose specific assumptions on the noise $\nu$, such as Gaussianity \citep{anderson1956statistical, lawley1962factor}, then a likelihood-based approach would be appropriate. However, in our setting we allow unknown, non-Gaussian $\nu$ for which there is no likelihood.
Then the question becomes, ``without a working distribution, how can one identify and estimate the model parameters?''. One could either use a quasi-likelihood approach \citep{bai2012statistical} and maximize some objective function, or one could use a PCA-based method on the observed covariance matrix $\widehat\Sigma_X$ to construct the factor loading space \citep{bai2003inferential, fan2013large}. Without additional assumptions on the latent structure, these methods can only estimate the \emph{column space} of $B$ (as opposed to $B$ itself). 
In contrast, our goal is to estimate not only the decoder matrix $B$ via the mixing model \eqref{eq.linmix}, but \emph{also} the latent structure through the latent causal model \eqref{eq.linSEM}. Thus, in order to simultaneously estimate both without a priori constraints, we require a fundamentally different approach.

By using multiple environments, the CRL literature provides an alternative to go beyond the aforementioned bottleneck and estimate the latent causal factors without distributional assumptions or constraints on the latent structure. Many of these results, starting with \citet{squires2023linear}, rely on the rank-one fingerprint left by $\Omega(d)$ single-node interventions in order to identify source nodes. Unfortunately, these techniques break down with multiple intervention targets: With multiple targets, the source nodes can be mixed with non-source nodes and the rank structure becomes more intricate.

To overcome this hurdle, instead of identifying source nodes from each environment \emph{separately}, we propose a novel technique that exploits \emph{combinations} of environments, and in doing so allows us to maximally extract identifying power from these environments. As a result, we need substantially fewer total environments, only $O(\log d)$ compared to $\Omega(d)$. Moreover, this approach involves only second-order statistics making it appropriate for general noise distributions. We introduce a pipeline that efficiently analyzes the intersections of column spaces of $\Sigma_X^{(k),\ell}$ over the different environments, extracts the dimensions of these intersections, and constructs the unknown intervention targets to then be used to identify the decoder $B$ and the latent causal graph $\gr$ via a generalized eigenvalue problem.

In this section, we present our basic methodology at the population level to simplify the initial exposition. 
In the next section (Section \ref{sec.finite.sample.guar}) we present our main error bounds and outline the challenges involved in extending these ideas to finite samples. 
We will proceed in three steps: (1) Consistently estimating the unknown multi-node intervention targets $I(k)$, (2) Consistently estimating the unknown linear mixing matrix $B$, and finally (3) Consistently estimating the latent graph, $\gr$. Note that from an estimate of $B$, we can directly estimate the causal representations $Z$ by taking $\widehat{Z}=\widehat{B}^{\dagger}X$.

\subsection{Reconstructing intervention targets}
Although not the main target of estimation, we first identify the unknown intervention targets. While this information is also useful in some applications, in our case it serves as an important ingredient for estimating the decoder $B$.

First, define the following quantity for any $T \subseteq [K]$:
\begin{align}
    g(T)
    ~:&=~ \dim\bigg[\bigcap_{k \in T}\col\Big(\Sigma_X^{(k),\ell}\Big)\bigg]. \label{eq.g}
\end{align}
This is the dimension of the shared column space between the environments indexed by $T$. The following lemma shows that these numbers suffice to identify the targets $I(k)$:
\begin{lemma}\label{lem.poly1}
The unknown intervention targets $I(k)$ can be identified from the collection $\{g(T) : T \subseteq [K]\}$. Moreover, there is a $\poly(d)$-time algorithm 
that recovers $I(k)$ from $\big\{\Sigma_X^{(k),\ell}\big\}_{k \in [K], ~\ell=1,2}$.
\end{lemma}
\noindent
It is clear that $g(T)$ is directly identified from the environment-specific covariances $\Sigma_X^{(k),\ell}$; computing $g(T)$ in practice amounts to routine SVD gymnastics. The main thrust of this lemma is that $I(k)$ can then be reconstructed from the $g(T)$, of which there are at most $O(d)$ to compute.

By exploiting rank information in the covariances in this way, we avoid distributional assumptions on the noise by focusing on second-order statistics. To see where this hidden rank structure comes from, 
we start with the environment-specific covariance matrices $\Sigma_X^{(k),\ell}$.
The column spaces of these matrices are directly linked to both the decoder $B$ and the intervention targets $I(k)$ through the following identity:
\begin{align}
    \col(\Sigma_X^{(k),\ell})=\col(B_{[\cdot,S(k)]}),
    \quad
    S(k) := I(k)^c. \label{eq.identity}
\end{align}
This suggests that the decoder and the intervention targets can be recovered through a careful analysis of the column spaces of the environment-specific covariances.

Of course, we don't know $B$ or the targets $I(k)$, but we can still exploit the hidden rank structure of $\Sigma_X^{(k),\ell}$, which are rank-deficient: With the exception of potential observational environments, $\rank(\Sigma_X^{(k),\ell})= |S(k)|:=r_k<d$. Moreover, each environment induces rank constraints on $B$: In the $k$th environment, only the columns in $B$ corresponding to $S(k)$ are effective in the model. The latter fact is exploited next to estimate the decoder $B$.

\begin{remark}
\label{rem:errordist}
    By relying solely on second-order statistics, we also avoid linear non-Gaussian (i.e. LiNGAM, \citealp{shimizu2006linear}) type assumptions that are commonly used for identifiability in CRL. Our results allow for arbitrary combinations of Gaussian and non-Gaussian latent variables and/or error terms.
\end{remark}

\subsection{Recovering the decoder} \label{sec.decoder}
The next step is to recover the decoder matrix $B$. Using the targets $I(k)$ recovered in the first step, define the set of environment indices that support node $j$ as follows:
\begin{align}
     \kappa_j := \big\{k: j \in  I(k)^c\big\} \label{eq.kappa.j}
\end{align}
\noindent
Then we take intersections of $\col(\Sigma_X^{(k),\ell})$ over $\kappa_j$ to recover each column $B_{\cdot j}$ one by one due to the following relation:
\begin{align}
    \col\big(B_{\cdot j}\big) = \bigcap_{k \in \kappa_j} \col\Big(\Sigma_X^{(k),\ell}\Big).\label{eq.lin.mix}
\end{align}
The reason $B$ can be recovered in this manner is because the column space of $\Sigma_X^{(k),\ell}$ spans the same space as the column space of $B_{[\cdot ,S(k)]}$ as shown in (\ref{eq.identity}). 
We can thus recover the entire mixing matrix (up to scale and permutation) and go beyond rotational invariance without having to assume the conventional (but involved) sparsity restrictions on $B$.

\subsection{Learning the latent causal graph} \label{sec.latent.graph}
Now that we have the decoder matrix $B$, it is straightforward to recover the representations $Z$ from $X$. The last step is to recover the latent causal graph $\gr$ over $Z$.

Using the recovered linear mixing $B$, we peel off the observed layer from the observational environment in $\Sigma_X^{(0),\ell} = B \Sigma_Z^{(0),\ell}B^T$ using the pseudoinverse $B^{\dagger}$ and solve a generalized eigenvalue problem for the now accessible latent covariances $\Sigma_Z^{(0),1}, \Sigma_Z^{(0),2}$:
\begin{align}
    \Sigma_Z^{(0),1}t ~&=~ \lambda\cdot\Sigma_Z^{(0),2}t. \label{eq.geq}
\end{align}
The generalized eigenvalue problem \eqref{eq.geq} yields $d$ solutions for the generalized eigenvectors $t_m$ ($m=1,\ldots,d$), and when combined into an $d \times d$ matrix, $T_Z=[\,t_1\given\cdots\given t_{d}\,]\in\R^{d\times d}$, we gain information on the latent causal graph $\gr$ through the following identity:
\begin{align}
  T_Z ~=~  \big(I - A^{(0)}\big) D \label{eq.gen.eigvec}
\end{align}
for some diagonal scaling matrix $D=(d_{jj})$ with $d_{jj}\ne0$.
Thus, $T_Z$ encodes the latent causal graph $\gr$ through its zero pattern, yielding a direct estimate of $\gr$ once we have the solutions to \eqref{eq.geq}.

\section{Statistical guarantees}\label{sec.finite.sample.guar}

In the previous section, we described a procedure at the population level for learning causal representations.
To extend this to a data-driven procedure that estimates causal representations from finite samples, a natural approach is to replace the environment-specific covariance matrices with their sample versions, i.e. 
\begin{align*}
    \widehat \Sigma_X^{(k),\ell} := \frac1n\sum_{i=1}^{n_{k,\ell}} \big[X^{(k),\ell}\big] \big[X^{(k),\ell}\big]^T
    \quad\text{ in place of }\quad
    \Sigma_X^{(k)\ell} = \E\Big(\big[X^{(k),\ell}\big] \big[X^{(k),\ell}\big]^T\Big).
\end{align*}
In this section we outline the resulting estimators and our main statistical results on the finite-sample behaviour of these estimators.

\subsection{Estimators}

In principle, the estimators can be straightforwardly derived by ``plugging-in'' sample covariances. However, in practice there is a thresholding step needed to handle finite-sample artifacts. Thus, before presenting the main statistical results, we formally define the estimators based on Section~\ref{sec.stat}.

We are given $n_{k,\ell}$ i.i.d. copies of $\{X_i^{(k),\ell}\}_{i=1}^{n_{k,\ell}}$ from model (\ref{eq.linSEM}-\ref{eq.linmix}) for each $k\in [K]$ and $\ell=1,2$. Denote the smallest sample size over all environments as $n_{\min}:=\min_{k,\ell}n_{k,\ell}$ and denote the maximum support size as $r:= \max_k r_k = \max_k |S(k)|$ where $S(k):=I(k)^c$ is the complement of the intervention target set as defined in (\ref{eq.identity}). 
For the finite-sample analysis, we will also assume for each $k,\ell$ that $X^{(k),\ell}$ is a centered sub-Gaussian random vector with $\norm{X^{(k),\ell}}_{\psi_2} \leq C_{\psi_2}$ for some finite constant $C_{\psi_2}>0$, with the sub-Gaussian norm defined in \citet{vershynin2018high}. To simplify the theorem statements, we also assume balanced sample sizes $n_{k,\ell}\asymp n/K$ although this is not needed in the proofs.

To facilitate stating the main results, we let the non-zero eigenvalues of $\Sigma_X^{(k),\ell}$ be $\lambda_1^{(k),\ell} \geq ... \geq \lambda_{r_k}^{(k),\ell} >0$ and define the following:\begin{align*} 
        \lambda^{-} ~&:=~ \min_{k,\ell}~\lambda_{r_k}^{(k),\ell},~~~~~
        \lambda^{+} ~:=~\max_{k,\ell}~\lambda_{1}^{(k),\ell}.
    \end{align*}

\paragraph{Intervention targets.}
In the first step of reconstructing $I(k)$ (cf. Section~\ref{sec.model}), we define a corresponding plug-in estimator for any $T \subseteq [K]$ similar to (\ref{eq.g}) in Section \ref{sec.stat}. Given a threshold $\rho \geq 0$ to be chosen later, let
\begin{align}
    \widehat{g}_{\rho}(T)
    ~&=~ \dim_{\rho}\bigg[\bigcap_{k \in T}\col\Big(\widehat\Sigma_X^{(k),\ell}\Big)\bigg] ~:=~ \Bigg|\bigg\{ i: \lambda_i\bigg(\Big[\underset{k \in T}{\prod} P_{\widehat\Sigma_X^{(k),\ell}}\Big]\Big[\underset{k \in T}{\prod} P_{\widehat\Sigma_X^{(k),\ell}}\Big]^T\bigg) \geq \rho\bigg\}\Bigg|, \label{eq.g.plugin}
\end{align}
where $\lambda_i$ denotes eigenvalues as mentioned in the notation section. Recall that $P_{\widehat\Sigma_X^{(k),\ell}}$ is the orthogonal projection onto $\col(\widehat\Sigma_X^{(k),\ell})$.
Thus, \eqref{eq.g.plugin} estimates the dimension of the shared column space between the environments indexed by $T$ by thresholding the eigenvalues of a certain matrix defined with respect to the sample covariance matrices $\widehat \Sigma_X^{(k),\ell}$. 
Of course, with population level inputs, the natural choice for tuning would have been $\rho=1$, but with finite samples determining the correct threshold is a more delicate matter.

\paragraph{Decoder matrix and representations.}
To estimate the decoder, no thresholding is needed: We simply use a vanilla plug-in version of \eqref{eq.lin.mix} (cf. Section~\ref{sec.decoder}):
\begin{align}
\widehat B_{[\cdot,~j]}~=~ \underset{k \in \widehat \kappa_j}{\bigcap} \col\big(\widehat{\Sigma}_X^{(k),\ell}\big) ~~~\text{where}~~~\widehat \kappa_j ~:=~ \big\{k: j \in \widehat I(k)^c\big\}. \label{eq.dec.sample}
\end{align}
From this, the representations $Z$ can be recovered simply by inversion, i.e. $Z=B^{\dag}X$.

\paragraph{Latent causal graph.}
Similar to how we constructed the $d \times d$ matrix $T_Z$ by stacking the $d$ generalized eigenvectors obtained from \eqref{eq.geq} (cf. Section~\ref{sec.latent.graph}), we define a corresponding plug-in quantity $\widehat T_Z$ by first solving an empirical generalized eigenvalue problem with $\widehat\Sigma_Z^{(0),\ell}$ instead of $\Sigma_Z^{(0),\ell}$, i.e.
\begin{align}
    \widehat\Sigma_Z^{(0),1} t ~&=~ \lambda\cdot\widehat\Sigma_Z^{(0),2} t. \label{eq.geq.sample}
\end{align}
As before, this yields $d$ generalized eigenvectors $\widehat{t}_m$, which we stack into the matrix $\widehat T_Z = [\widehat{t}_1\given\cdots\given\widehat{t}_d]$.
We then define an estimator of $\gr$ as the graph $\widehat{\gr}_{\alpha}$ obtained by thresholding $\widehat T_Z$ with a tuning parameter $\alpha$; i.e.
whose edges are determined by
\begin{align}
    \text{edge } j_1\to j_2 \text{ is in }\widehat{\gr}_{\alpha}~\Longleftrightarrow~ \big|[\widehat{T}_{Z}]_{[j_1,j_2]}\big| > \alpha. \label{eq.lcg.sample}
\end{align}
Again, determining a data-driven level for $\alpha$ that correctly recovers $\gr$ is key.

\paragraph{Main result.}
Before presenting the main theorems, we present an informal summary of the main results in the following:
\begin{thm}[Informal] \label{thm.sum.est}
    Under \ref{assum.PSSS.0}-\ref{assum.rho}, bounded eigenvalues for $\Sigma_Z$, and balanced samples, we have with probability greater than $1-(1/pn)$:
    \begin{align}
\underset{\diagonal~D_d\succ0}{\inf}~\frac{1}{\sqrt{d}}\norm{\widehat B - B D_d}_F ~&\lesssim~  \sqrt{\frac{r \log (pn)}{n}}.
\end{align} 
    Moreover, with the same probability, $\widehat{\gr}_{\alpha} = \gr$ for $\alpha  ~\asymp~ \sqrt{\dimz^2 \log (pn)/n}$ as long as
    \begin{align}
    a_{\min} 
    ~&\gtrsim~ \sqrt{\frac{\dimz^2 \log (pn) }{n}}. 
    \end{align}
\end{thm}
\noindent
While implementation with (\ref{eq.g.plugin}-\ref{eq.lcg.sample}) is straightforward in practice, deriving the finite-sample error bounds above for such a ``plug-in'' scheme presents several challenges:
\begin{enumerate}
\item As in both classical and high-dimensional factor analysis, we only have noisy views of the column spaces based on the empirical covariances $\widehat\Sigma_X^{(k),\ell}$. Although tools like Davis-Kahan and Wedin- or Weyl-type bounds are available, $g(T)$ involves taking intersections over multiple environments, which complicates the application of these bounds, considering that $\col(\widehat W_1) \cap \col(\widehat W_2)$ is a non-linear transformation of $(\widehat W_1, \widehat W_2)$.
\item Learning the latent causal graph $\widehat \gr (A)$ requires a perturbation analysis on generalized eigenvectors, which involves the inversion of empirical covariances $\widehat \Sigma_Z^{-1/2}$ and empirical pseudoinverses $\widehat B^{\dagger}$ that are ill-conditioned and thus present additional challenges compared to standard eigenvalue perturbation bounds.
\end{enumerate}

\noindent
In the sequel, we describe how these challenges can be overcome.

\subsection{A key tool: Projection based eigen-counting}\label{sec.technical.device}

A key step in the analysis is to understand the sensitivity of the discrete object $g(T)$ to noisy perturbations of the column spaces. In order to gain control over this, we will relate $g(T)$ to the spectral behaviour of certain projection matrices.

Lemma~\ref{lem.proj} below presents a general result that connects a carefully constructed projection matrix to the intersection of linear spaces that is essential to controlling deviations in $g(T)$. Define
\begin{align}
    Q(T) ~:=~ \bigg(\prod_{k \in T}P_{\Sigma_X^{(k),\ell}}\bigg) \bigg(\prod_{k \in T}P_{\Sigma_X^{(k),\ell}}\bigg)^T \label{eq.Q}
\end{align}
To keep $Q(T)$ well-defined, we order these projections by ascending $k$. We first relate the eigenspace of $Q(T)$ to the shared column space that defines $g(T)$ in \eqref{eq.g}.
\begin{lemma} \label{lem.proj}
     Consider the symmetric square matrix $Q(T)$ defined in \eqref{eq.Q}. The eigenspace of $Q(T)$ spanned by the unit eigenvalues, denoted by $\Eig_1(Q(T))$, exactly equals the shared column space across $T$, i.e. $$\Eig_1 (Q(T))~=~ \bigcap_{ k \in T}\col(\Sigma_X^{(k),\ell}).$$
     In particular, we have
     \begin{align*}
         \dim\Bigg(\bigcap_{ k \in T}\col(\Sigma_X^{(k),\ell})\Bigg) ~=~ g(T) ~=~ \rank\big(\Eig_1(Q(T))\big).
     \end{align*}
\end{lemma}
\noindent
When applied to the column space of the observed covariance matrices, this result establishes the second equivalence in (\ref{display.equivalence.0}). Thus, we so far have the following useful equivalences:
\begin{align}
    I(k),~ k \in [K] ~~~\overset{\text{Lemma }3.1}{\Longleftrightarrow}~~~
     g(T),~ T \subseteq [K] ~~~\overset{\text{Lemma }4.1}{\Longleftrightarrow}~~~ \rank\big(\Eig_1(Q(T))\big),~ T \subseteq [K].   \label{display.equivalence.0}
\end{align}
This suggests that deviations in $g(T)$ are dictated by the spectral properties of $Q$, and provides a convenient link to control deviations in $g(T)$ through $Q(T)$.
For the resulting perturbation analysis, we need an empirical counterpart of \eqref{eq.Q}, $\widehat{Q}$ using sample covariances:
\begin{align}
    \widehat{Q}(T) ~:=~ \bigg(\prod_{k \in T}P_{\widehat\Sigma_X^{(k),\ell}}\bigg) \bigg(\prod_{k \in T}P_{\widehat\Sigma_X^{(k),\ell}}\bigg)^T \label{eq.Q.hat}
\end{align}
\noindent
The next lemma is a general perturbation bound on $Q(T)$ that makes this link explicit:
\begin{lemma}\label{lem.perturbation}
For all non-empty subsets of intervention indices $T\subseteq [K]$, with $Q(T),~\widehat Q(T)$ defined in (\ref{eq.Q}) and (\ref{eq.Q.hat}), respectively, we have for all eigenvalues $\lambda_j(\cdot),~1\leq j \leq p$, the following uniform perturbation bound with probability $\geq 1-(1/pn)$:
    \begin{align}
    \underset{T \subseteq [K]}{\max}~\underset{1\leq j \leq p}{\max}~ \bigg|\lambda_j\Big( \widehat{Q}(T)\Big) - \lambda_j\Big( Q(T)\Big)\bigg| 
    ~\lesssim~ \frac{\lambda^+}{\lambda^-}\cdot\sqrt{\frac{r \log (pn)}{n}}\label{eq.weyl.P}
\end{align}
\end{lemma}

Now that we have control over perturbations to $Q(T)$, the last step is to connect this back to $g(T)$. First, by \eqref{eq.g.plugin}, we have
\begin{align*}
    \widehat{g}_\rho(T)
    =  \Big|\big\{ i: \lambda_i\big(\widehat{Q}(T)\big) \geq \rho\big\}\Big| .
\end{align*}
When the matrix $B$ is severely ill-conditioned, the observed covariances $\Sigma_X^{(k),\ell}$ become ill-conditioned. If this is the case, then $\widehat Q(T)$ is no longer reliable to recover $g(T)$. To avoid this kind of degeneracy, we introduce a mild regularity condition on the eigengap in $B$:

\begin{enumerate}[label=(A\arabic*),start=3]
    \item\label{assum.rho} The decoder $B\in\R^{p\times d}$ satisfies
    \begin{align*}
    \frac{\lambda_{\min}(B^TB)}{\lambda_{\max}(B^TB)} := 1-\rho^* ~\gtrsim~ \Bigg\{ \bigg(\frac{\lambda^+}{\lambda^-}\bigg)^2\cdot\frac{\dimz r \log (pn)}{n}\Bigg\}^{1/3},
    \end{align*}
    for $\rho^*=\rho^*(B)$ defined above.
\end{enumerate}
Thus, while we allow for ill-conditioned $B$ (i.e. $1-\rho^*\approx 0$), the severity of this ill-conditioning is controlled by \ref{assum.rho}. In reasonable settings with $p\gg n \gg \dimz$, this lower bound is allowed to vanish as $n\to\infty$ (see Remark~\ref{rem.assum.rho}).

\begin{lemma}\label{lem.perturbation.2}
    Under \ref{assum.rho}, we have 
 \begin{align*}
        \mathbb{P}\Big[\widehat{g}_{\rho }\big( T\big) = g\big( T\big) \FORALL T \subseteq \{1,2,...,K\}\Big]~\geq~1-\frac{1}{pn}
    \end{align*}
    for the estimand $g(T)$ defined in (\ref{eq.g}) and the plug-in estimator $\widehat g_{\rho}(T)$ defined in (\ref{eq.g.plugin}) with tuning parameter $\rho$ satisfying $1-\rho \les \frac{\lambda^+}{\lambda^-}\sqrt{r \log (pn)/n}$.
\end{lemma}
\noindent
Lemma \ref{lem.perturbation} is the perturbation result that uniformly controls the deviation of all eigenvalues of $Q(T)$ over all possible environment index sets $T$; Lemma \ref{lem.perturbation.2} shows that with a data-driven choice of $\rho$, accurate eigen-counting is guaranteed (with high probability) even with noisy versions of the column spaces. This result connects the equivalence structure in (\ref{display.equivalence.0}) to our finite-sample estimators.

\begin{remark}
\label{rem.assum.rho}
    Condition~\ref{assum.rho} is closely related to existing assumptions from the literature, and in many cases significantly relaxes these assumptions. For example, the commonly assumed pervasiveness assumption \citep{fan2008high, fan2013large} requires $1-\rho^*\asymp 1$, which obviously implies \ref{assum.rho}. Another example is when the entries of $B$ are i.i.d. sub-Gaussian: In this case, $1-\rho^*\asymp 1$ with high probability. More generally, \ref{assum.rho} holds trivially when the columns of $B$ are nearly orthogonal, which of course holds with high probability in the i.i.d. sub-Gaussian setting. By contrast, we allow $1-\rho^*$ to vanish asymptotically, allowing for $B$ to become ill-conditioned in high-dimensional settings. To emphasize this, we leave the dependence on $\rho^*$ explicit in our results, noting that under any of the above settings, this will just be constant.
\end{remark}

\subsection{Finite-sample rates}

The pipeline for converting noisy column spaces into accurate estimates of $g(T)$, combined with Lemma~\ref{lem.poly1}, ensures high probability recovery of the intervention targets $I(k)$:
\begin{thm}\label{lem.S_k.0}
    $\mathrm{(Intervention~ target ~estimation~consistency)}$ Under \ref{assum.rho}, with the plug-in estimator in (\ref{eq.g.plugin}) with tuning parameter $1-\rho \les \frac{\lambda^+}{\lambda^-}\sqrt{r \log (pn)/n}$, the estimated intervention sets $\widehat{I}(k)\FORALL k$ satisfy
    \begin{align*}
         \mathbb{P}\Big[\widehat I(k)=I(k)\FORALL k\in[K]\Big]~\geq~ 1-\frac{1}{pn}.
    \end{align*}
\end{thm}

Next we present the second main result, estimation consistency for the mixing matrix, $B$. 

\begin{thm} \label{thm.G.0.fixed} $\mathrm{(Mixing~matrix~estimation~consistency;~Fixed~design)}$
Under \ref{assum.PSSS.0} and \ref{assum.rho}, we have with probability greater than $1-(1/pn)$ that
\begin{align*}
    \underset{\diagonal~D_{\dimz}\succ0}{\inf}~\norm{\widehat B - B D_{\dimz}}_F 
    ~&\lesssim~
    \frac{\sqrt{\dimz}}{1-\rho^*}\cdot\frac{\lambda^+}{\lambda^-}\sqrt{\frac{r \log (pn)}{n}}
\end{align*}
\end{thm}
\noindent
Since each column in $B$ is only identifiable up to scale, the result is presented in terms of an infimum over all possible positive definite diagonal matrices $D_{\dimz} \in \R^{\dimz \times \dimz}$.

Finally, we present a finite-sample result for estimating the latent causal graph. To simplify the main result, we state the result here assuming $\Sigma_Z^{(0),1}, \Sigma_Z^{(0),2}$ have bounded eigenvalues; in our technical results the eigenvalues of $\Sigma_Z$ are allowed to be general. For notational convenience we express the above by simply saying $\Sigma_Z$ has bounded eigenvalues.
Let $a_{\min}$, $a_{\max}$ denote the minimum and maximum non-zero coefficients in $A$, and $m_{\deg}$ denote the maximum in-degree of $\gr$.
\begin{thm} \label{thm.B.fixed}
    $\mathrm{(Latent~graph~estimation~consistency;~Fixed~design)}$
    Under \ref{assum.PSSS.0}-\ref{assum.rho}, and $\lambda_{\min}(\Sigma_Z)\asymp\lambda_{\max}(\Sigma_Z)\asymp1$, 
we have \begin{align}
        \pr\big[\widehat{\gr}_{\alpha} = \gr\big] ~\geq~ 1-\frac{1}{pn} 
        \quad\text{with}\quad
        \alpha ~\asymp~ \frac{1}{1-\rho^*}\cdot\bigg(\frac{\lambda^+}{\lambda^-}\bigg)^3\cdot\sqrt{\frac{\dimz^2 \log (pn)  }{n}} ,
        \label{eq.prob.fixed}
    \end{align} 
    given the following holds:
\begin{align}
    a_{\min} 
    ~&\gtrsim~ \frac{a_{\max} \sqrt{m_{\deg}}}{1-\rho^*}\cdot\bigg(\frac{\lambda^+}{\lambda^-}\bigg)^3\cdot\sqrt{\frac{\dimz^2 \log (pn) }{n}}.
    \end{align}
\end{thm}

\noindent
Recall that $\wh{\gr}_\alpha$ is defined by thresholding the estimated generalized eigenvectors $\widehat {T}_Z^{(k)}$.
To determine a precise statistical rate for $\alpha$, we undergo a perturbation analysis on the generalized eigenvectors.

\begin{remark}
    Bounded eigenvalues for $\Sigma_Z$ is a common assumption \citep{fan2011high,park2020identifiability,bing2022inference}, and thus motivates our presentation of the rate in Theorem \ref{thm.B.fixed}. 
    Note that this boundedness assumption can be made regardless of whether pervasiveness (i.e. $\sigma^2_{\min}(B) \asymp \sigma^2_{\max}(B) \asymp p$) is assumed \citep{fan2011high,fan2013large,wang2017asymptotics} or not. 
    Our technical results include the case where the largest eigenvalue of $\Sigma_Z$ grows with the latent dimension $d$. 
\end{remark}

\section{Discussion}
This paper studies causal representation learning in linear latent factor models in high-dimensions with multiple unknown intervention environments. 
Our main contribution is to provide finite-sample guarantees for a novel estimator of causal representations that applies to settings with only a logarithmic number of environments. 
This paper helps fill a gap between recent identifiability theory in CRL and finite sample estimation questions that come up in practice, while also matching the known lower bound for the number of environments. 

The crux of the method is that the latent structure can be extracted from second-order statistics through careful exploitation of the information in different combinations of environments. The central technical device, a projection-based eigen-counting procedure, converts noisy estimates of column space intersections into consistent recovery of the unknown intervention targets, $I(k)$, which is then used to isolate each column in the decoder $B$, which in turn leads to a generalized eigenvector analysis on the recovered latent covariances for consistent estimation of the latent graph, $\gr$. 

We hope these results provide a useful step toward a statistical understanding of causal representation learning in which identifiability, estimation, and intervention design are all analyzed within a common framework. Of course, our results raise more questions than they answer, and this is just a first step in this endeavour.

\bibliographystyle{abbrvnat}
\bibliography{ref.bib}

\end{document}